# Navigation beyond Wayfinding: Robots Collaborating with Visually Impaired Users for Environmental Interactions

Shaojun Cai
shaojun@u.nus.edu
National University of Singapore
Singapore  Singapore

Nuwan Janaka
nuwanj@u.nus.edu
City University of Hong Kong
Hong Kong  China

Ashwin Ram
ram@cs.uni-saarland.de
Saarland University
Saarbrücken  Germany

Janidu Shehan
shehanmkj.21@uom.lk
University Of Moratuwa
Moratuwa  Sri Lanka

Yingjia Wan[*]
wanyj@psych.ac.cn
Institute of Psychology, Chinese Academy of Sciences
Beijing  China

Kotaro Hara
kotarohara@smu.edu.sg
Singapore Management University
Singapore  Singapore

David Hsu[*]
dyhsu@comp.nus.edu.sg
National University of Singapore
Singapore  Singapore

**Existing Systems**

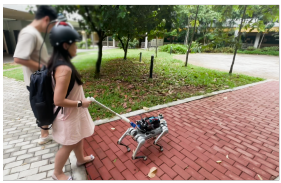

**Wayfinding**

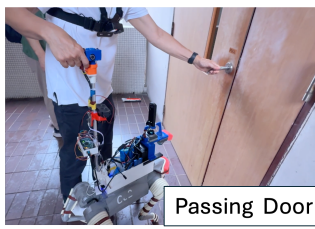

Passing Door

**Our System**

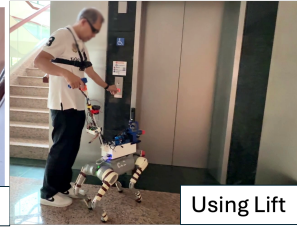

Using Lift

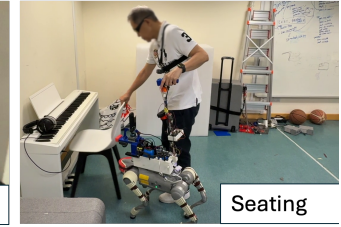

Seating

**Environmental Interactions**

Figure 1: Existing robotic guide systems mainly support wayfinding (left). Our proposed system enables environmental interactions, including door access, elevator use, and seating (right).

## Abstract

Robotic guidance systems have shown promise in supporting blind and visually impaired (BVI) individuals with wayfinding and obstacle avoidance. However, most existing systems assume a clear path and do not support a critical aspect of navigation—*environmental interactions* that require manipulating objects to enable movement. These interactions are challenging for a human–robot pair because they demand (i) precise localization and manipulation of interaction targets (e.g., pressing elevator buttons) and (ii) dynamic coordination between the user's and robot's movements (e.g., pulling out a chair to sit). We present a collaborative human–robot approach that combines our robotic guide dog's precise sensing and localization capabilities with the user's ability to perform physical manipulation. The system alternates between two modes: **lead mode**, where the robot detects and guides the user to the target, and **adaptation mode**, where the robot adjusts its motion as the user interacts with the environment (e.g., opening a door). Evaluation results show that our system enables navigation that is safer, smoother, and more efficient than both a traditional white cane and a non-adaptive guiding system, with the performance gap widening as tasks demand higher precision in locating interaction targets. These findings highlight the promise of human–robot collaboration in advancing assistive technologies toward more generalizable and realistic navigation support.

## 1 Introduction

Worldwide, an estimated 290 million people are blind or visually impaired (BVI), including 43 million who are completely blind [1]. For many, independent navigation—particularly in unfamiliar or complex environments—remains a major barrier to autonomy and social participation [27]. White canes are widely used due to their portability and ease of use, but are limited to short-range environmental and obstacle perception.

Robotic guidance systems have made substantial progress in supporting independent travel for BVI individuals [7, 13, 27, 38]. They provide intuitive and reliable navigation assistance, proven to be effective and user-friendly, particularly in complex and unstructured environments such as cafeterias [7], museums [16], and other public spaces [35]. However, most existing systems primarily focus on leading users along a clear path—following predefined routes, avoiding obstacles, and stopping at destinations [13, 19, 22, 39]. This narrow focus overlooks a critical aspect of everyday navigation—*environmental interactions*, tasks that require users to locate, approach, and manipulate small or visually subtle objects, such as pressing elevator buttons, opening doors, or pulling out chairs. Without assistance, these interactions are often stressful, time-consuming, and socially awkward.

These environmental interactions introduce challenges that require task-aware coordination rather than simple guidance. Humans and robots possess complementary but asymmetric capabilities: humans can manipulate objects but lack visual access, whereas robots can perceive the environment, but performing fine-grained physical interactions remains complex and costly. Most existing systems fail to bridge this asymmetry, stopping near the target and leaving users to locate visually subtle interaction points, such as buttons or handles—often through trial and error—which places a disproportionate cognitive and physical burden on the user.

To address these challenges, we first conducted interviews and prototype trials with six BVI participants, identifying key design requirements: accurate positioning, accessible and adaptive placement, and clear multi-modal communication. Building on these insights, we propose a collaborative human–robot approach that combines the robotic guide dog's precise sensing and localization capabilities with the user's ability to physically manipulate objects. At its core is a **dual-mode collaboration system** in which the robot first detects and guides the user toward the interaction target (*Lead mode*), and then dynamically adapts its motion as the user physically interacts with the environment (e.g., opening a door), maintaining proximity without interference (*Adaptation mode*).

We evaluate this system in three representative scenarios—elevator use, door operation, and seating—that capture the diversity of real-world navigation. Through controlled studies and route-level field trials, we demonstrate that our approach improves efficiency, safety, and user experience compared to traditional aids and a non-adaptive baseline system. Our contributions are threefold:

- We present a robotic guide system that addresses a critical gap in existing assistive technologies by supporting *environmental interactions* during navigation—tasks that require precise, task-aware guidance beyond simple path-following.
- We propose a dual-mode human–robot collaboration approach that combines precise robotic sensing with adaptive user-guided interaction, with experiments demonstrating improvements over a non-adaptive baseline.
- We validate this approach in real-world route-level navigation tasks, showing improvements in efficiency, safety, and user experience over non-adaptive robotic systems and traditional mobility aids (white cane and guide dog).

## 2 Related Works

### 2.1 Assistive Robotics and Guide Systems

A long line of research has developed robotic guide systems for BVI individuals [13, 27, 38]. with early prototypes focusing on path-following and obstacle avoidance [22, 24, 33, 37] and more recent systems demonstrating reliable assistance in complex public spaces such as museums and cafeterias [7, 13, 19]. Despite their effectiveness in wayfinding, these systems are typically designed for users to passively follow a predefined path and offer limited support for environmental interaction tasks during navigation. In a recent field trial [35], participants frequently reported difficulties using elevators and locating call buttons from a distant position, often requiring external assistance even when guided by autonomous robots. This highlights a key gap: while prior systems succeed in getting users near a destination, they rarely support the precise and task-aware interactions required once there.

Outside robotics, several assistive technologies have been developed to support object finding. Mobile applications such as *Be My Eyes* [5] and *Seeing AI* [3] provide object recognition or remote assistance through smartphones. while smart glasses [2, 4, 26, 43] use computer vision to identify objects. However, these tools primarily focus on environmental description rather than providing accurate, point-to-point guidance.

### 2.2 Object-Goal Navigation

In parallel, research in embodied AI has advanced object-goal navigation (OGN), where a robot autonomously navigates to an object specified by category label (e.g., "chair," "elevator") [4, 8, 20, 28]. These methods combine semantic perception with navigation policies, and are often trained in large-scale simulators such as Habitat [34]. While relevant to our environmental interaction tasks in approaching the object, most OGN methods treat navigation as complete once the robot reaches the vicinity of the object. For our tasks, however, arriving near the object is insufficient: success depends on guiding the user to a precise interaction point (e.g., an elevator button or chair handle) and supporting subsequent manipulation. Although some systems extend OGN with robotic manipulators [18, 42], such approaches remain limited by manipulation complexity, are often confined to controlled lab settings, and typically lack coordination with a human partner.

### 2.3 Mixed-Initiative Control

Another relevant line of work in HRI investigates mixed-initiative control, emphasizing fluid role negotiation between humans and robots. Studies show that flexible role switching improves fluency and reduces workload in collaborative tasks such as assembly and cooperative manipulation [21, 25, 36].

In robotic guide systems, recent shared-control approaches allow users to influence navigation decisions but largely focus on path-following and high-level choices such as turns at intersections [19, 29]. Related work has also examined blind users' preferences across autonomy levels in social navigation contexts [15], which is complementary but distinct from our focus on environmental interactions.

To summarize, prior mixed-initiative methods support reaching a general destination but rarely enable task-aware environmental interactions. To address this key gap, we introduce a dual-mode collaboration approach, where the robot guides the user to the target and adaptively yields initiative to support the user's interaction.

## 3 Design Requirement Exploration

*Overview.* To derive requirements for a robotic guidance system that addresses environmental interactions and enhances BVI users' experience, we conducted a formative study (Sec. 3) aimed at identifying the challenges in these interactions, analyzing the steps involved, and deriving design requirements for a system capable of overcoming them. Based on participant feedback, we iteratively designed the robotic guide dog (Sec. 4) and subsequently evaluated the final system in real-world settings through a user study (Sec. 5). All studies were approved by our Institutional Review Board (IRB).

We first conducted semi-structured interviews (**??**) with six BVI individuals experienced in using various assistive tools for navigation. Participants consistently reported that interactions requiring *locating and operating small, specific elements* were the most challenging aspect of independent travel. For example, one participant noted, "*Elevators are normally in relatively secluded areas of the building. But afterwards, you still need to locate the button, which is more challenging.*" Finding empty seats was similarly difficult due to unstructured layouts, clutter, and varying chair designs. Some participants had tried technology-based solutions, such as AI object-recognition apps (e.g., SeeingAI [3]) or smart glasses (e.g., Meta Rayban Glass [2]), but these tools only detect what the camera is already pointing at, failing to address the fundamental challenge of correctly aiming at non-obvious or unpredictable targets.

We then conducted a technology-probing [6] session with a visually impaired staff member from a local nonprofit mobility organization, who tested our prototype in his office building. A key insight was the importance of *accurate pointing to reduce unnecessary hand movements*. For example, when entering a room, he spent about 30 seconds locating the door handle, noting that "*the robot should point directly to the button—within 5 cm of the handle—accompanied by voice instructions specifying direction and height.*" We tested two guiding mechanisms: direct head pointing and voice-based clock-direction instructions , and the former was more effective. He also suggested combining direct head pointing with additional voice cues about the target's vertical position, so that he can more precisely locate the target on the first attempt.

Another insight concerned the *robot's adaptation strategy* during user–environment interactions. For example, when waiting for an elevator, the participant preferred the robot to position itself at a 45-degree angle to the door rather than directly in front, to avoid blocking people exiting. He also emphasized the importance of clear, efficient movements, noting that excessive adjustments during the adaptation stage could be distracting. Lastly, he emphasized the need for the ability to start and stop the robot at will, to maintain a sense of control and autonomy.

*Design Requirements.* Based on the interviews and on-site study, we derived the following design requirements.

- **D1**: **Accurate and Task-aware Positioning.** The robot should reliably position the user to minimize search effort and allow natural access to small interaction targets (e.g., handles, buttons). Guidance should be intuitive and precise, combining physical cues with complementary voice instructions as needed.
- **D2**: **Adaptive to User Interactions.** The robot should adjust its placement and behavior to avoid obstructing the user or environment, maintaining proximity that is supportive but non-intrusive. This allows smooth and safe execution of tasks such as opening doors or pulling out chairs.
- **D3**: **Effective Communication.** The robot and user should engage in clear, low-burden two-way communication to coordinate actions. The robot should provide transparent cues about its intentions, while allowing the user to easily confirm, initiate role switching, or override actions through lightweight mechanisms (e.g., button interface), ensuring shared control without unnecessary effort.

## 4 System Design

Guided by the requirements (section 3), our system integrates hardware, perception, planning, and interaction interface to address accurate target localization (**D1**), non-obtrusive positioning for user manipulation (**D2**), and multimodal communication (**D3**). We first describe the hardware platform, including sensors, leash interface, and input controls that support **D1** and **D3**. We then present the perception pipeline for robust detection of main objects and interaction targets (**D1**). Next, we introduce the dual-mode planner, which alternates between *lead mode*—precisely positioning the user without blocking their workspace (**D2**)—and *adaptation mode*, where the robot yields initiative during interaction. Finally, we outline the user interfaces enabling confirmation, role-switching, and communication (**D3**). Figure 2 summarizes the overall architecture.

### 4.1 Hardware System

To support adaptation to different terrains in both indoor and outdoor environments, we adopted a low-cost Unitree Go2 quadruped robot [32], equipped with an onboard Jetson Orin NX computer [12] for planning and state estimation. We additionally installed a Jetson Orin AGX [11] to handle real-time processing of computer vision modules.

**Omni-Directional Perception**. We equipped the robot with a Ricoh Theta Z1 360° camera that provides omnidirectional coverage to support environmental perception and enable close-range object tracking (e.g., door handles, user hands, and user pose), as well as a Mid360 LiDAR sensor that captures accurate surrounding depth information (Figure 3). Panoramic images are modeled using equirectangular projection, and the extrinsic transformations between the LiDAR and the camera are computed using the method of [41]. Due to the sparsity of LiDAR scans relative to image pixels, bilinear interpolation is used to estimate intermediate values.

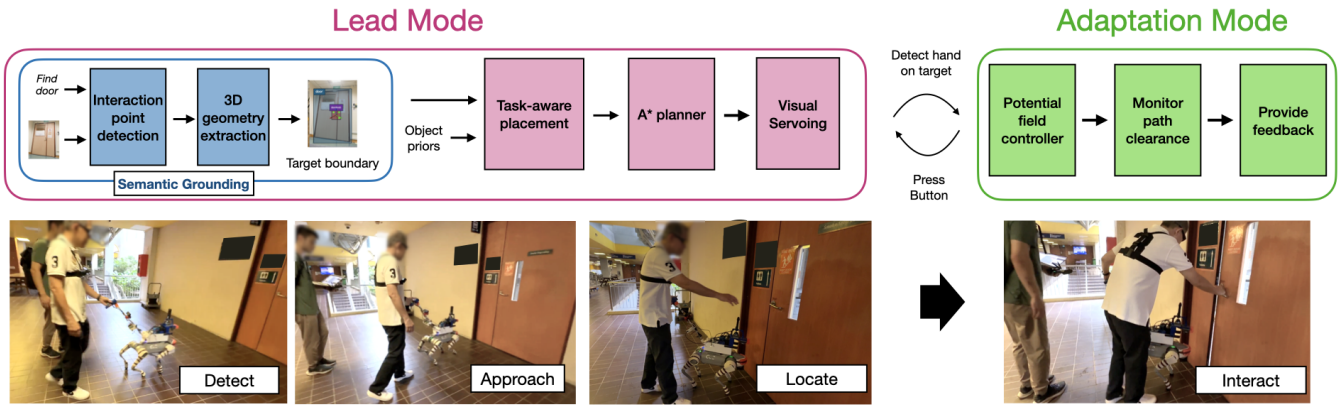

Figure 2: Overview of the dual-mode human-robot collaboration system. The system operates in two modes: *Lead Mode*, which involves semantic grounding, task-aware placement, planning, and visual servoing for guiding toward interaction points; and *Adaptation Mode*, which involves path monitoring and feedback for safe interaction.

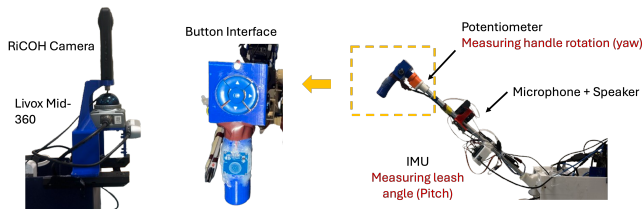

Figure 3: System apparatus. Left: Ricoh Theta Z1 camera [31] and Livox Mid360 LiDAR [10] setup. Middle: Button interface. Right: Leash design with IMU setup to measure leash angle (pitch) and handle rotation (yaw).

**Leash and Handle Design**. For precise guidance (**D1**), we allow longitudinal freedom along the robot's body axis so that users can push the leash forward to trace the robot's orientation and head position. Paired with voice instruction on the target height, the user can accurately locate the interaction point (**D3**). We also estimated the leash state with an IMU installed on the handle to classify whether it has been pushed into the pointing state; in adaptation mode, this generates a forward attraction vector indicating the user's intent to move forward (**D2**). Finally, a button interface was added, featuring an accessible 5-way directional button to support additional interactions between human and robot (**D3**).

### 4.2 Semantic and Geometry Perception

To support environmental interactions, the system must recognize and characterize relevant objects in the environment. We define the *interaction target* as the element the user interacts on (e.g., door handle, call button, or seat back), and the *main object* as the object whose state changes as a result of the interaction (e.g., a door, elevator door, or chair). We combine 2D detection with 3D geometry estimation to compute accurate object positions and relationships.

**2D Detection**. We employ NanoOWL open-vocabulary detector [23] and Segment Anything 2 [30] to extract the masks of the main objects and their interaction target. Those masks are then tracked in future frames using XMem [9]. In addition, we also extract the bounding box of the categories *person* and *hand* from NanoOWL and use them in the adaptation mode in Sec 4.3.3.

**3D Geometry Feature Extraction** After obtaining the mask, we perform contour detection on the main object, associate the contour points with 3D LiDAR data, and compute the object's width and height in 3D. Similarly, we compute the 3D center point of the interaction target. We also estimate the normal vector of the interaction target by sampling a planar region near the target center.

These geometric features are used to improve detection reliability by filtering candidates against predefined 3D priors (e.g., object width, height, and target height), retaining only detections within valid thresholds and selecting the highest-confidence candidate. The resulting spatial relationship between the interaction target and the main object guides user positioning (Sec. 4.3), while the target height above the ground is conveyed via voice feedback to support efficient object locating.

### 4.3 Dual-Mode Collaboration System

One key design of our system is the dual-mode human-robot collaboration. The robot switches between the **lead mode**, in which it guides the user, and the **adaptation mode**, in which it adapts to the user's movement and command. Below, we describe the role-switching mechanism and the design of the two modes.

*4.3.1 Role-switching mechanism.* The robot starts in **lead mode**, guiding the user from the starting position to the main object and positioning toward the interaction target while announcing its location. Upon detecting the user's hand on the interaction target, the robot switches to **adaptation mode**, where the user performs the manipulation and the robot adapts to the user's motion (e.g., opening a door), yielding to obstacles and preparing for subsequent navigation. Once the path is clear and the user confirms via button input, the robot switches back to **lead mode** to continue guiding.

*4.3.2 Lead Mode Implementation.* The objective of the lead mode is to help the user efficiently access the interaction target, and then resume leading the user once the interaction is finished. The three main components are listed below.

**Task-Aware Placement**. Firstly, the system computes two goals: a *stop goal* **x** for the user to reach and interact with the target, and an *end goal* for the task (e.g., the other side of a door or inside an elevator). The system also prompts the user to approach the target from the appropriate side. The stop goal is determined by three criteria: (1) collision avoidance, (2) ensuring the user can reach the target without releasing the handle, and (3) avoiding interference with the main object's motion. We formulate this as:

$$\mathbf{x}^* = \arg\min_{\mathbf{x}} \ w_c\, C(\mathbf{x}) \ + \ w_d \|\mathbf{x} - \mathbf{x}_t\| \ + \ w_m\, M(\mathbf{x}) \qquad (1)$$

Here, $C(\mathbf{x})$ penalizes collisions, $\|\mathbf{x} - \mathbf{x}_t\|$ enforces proximity to the interaction target $\mathbf{x_t}$, and $M(\mathbf{x})$ penalizes poses that fall inside the region swept by the object's motion (e.g., door swing area or chair pull-out range). The obstruction cost is estimated from the physical boundary of the main object and the estimated movement from prior knowledge. This encourages the robot to stop at collision-free, reachable locations that minimize interference with object motion. The weights $w_c, w_d, w_m$ balance safety, reachability, and task smoothness.

The *end goal* is computed based on task context: for doors, it lies on the opposite side of the doorway to guide the user through; for elevators, it is located inside the cabin; and for seats, it corresponds to the free space in front of the chair once pulled out. Together, the stop and end goals allow the robot to facilitate the interaction and ensure a smooth continuation of the navigation task.

**User Side Positioning**. In addition, the robot also determines which side of the robot the user should approach the interaction target from. For doors and chairs, the standing side is the side closer to the center of the main object, so that user's arm does not obstruct the robot's path after interaction is done. For elevator, the standing side is further from the main object, allowing the user to hold the elevator call button while following the robot into the elevator. The robot prompts the user of the standing side through a voice, "*please stand on the left / right side of the robot*".

**Accurate Positioning to Interaction Target**. We employ A* planner to calculate a collision free path to the stop goal calculated above, and estimate the robot's pose and surrounding obstacles using Fast-LIO [40]. To further reduce the accumulated positioning error, we perform visual servoing to accurately align the robot's head towards the interaction target. Our visual servoing module combines image-based and position-based strategies. The system first applies image-based visual servoing (IBVS) to align the target mask with the image center, rotating until the target falls within a central threshold. It then applies a position-based control using LiDAR depth to maintain a comfortable interaction distance (0.3–0.8 m). This hybrid approach reduces residual errors in both angle and distance (**D1**).

*4.3.3 Adaptation Mode Implementation.* Once the robot detects the user's hand on the interaction target, it switches to adaptation mode to accommodate the interaction.

**Potential Field Controller.** We employ a modified Artificial Potential Field (APF) controller [17]. The overall control force is given by:

$$\mathbf{F} = k_{\text{att}}\bigl((\mathbf{x}_u - \mathbf{x}) + \alpha \mathbf{h}\bigr) + \sum_i \mathbf{F}_{\text{rep}}(d_i) \qquad (2)$$

Here, **x** is the robot position (2D), $k_{\text{att}}$ scales the attraction term, and $\alpha$ scales the handle input. The repulsive force $\mathbf{F}_{\text{rep}}(d_i)$ is computed from the projected 2D positions of obstacles detected by the 3D LiDAR on the navigation plane at distance $d_i$. The attractive force is derived from the sensed user position ($\mathbf{x}_u$), complemented by the interactive handle signal (**h**), both represented as 2D vectors in the robot's local frame.

The robot orientation is set as a weighted blend of handle rotation $\theta_h$ (from the potentiometer) and the angle toward the end goal $\theta_g$:

$$\theta = \beta \theta_h + (1 - \beta)\, \theta_g, \qquad (3)$$

with $\beta \in [0, 1]$ balancing user input and goal alignment.

**Monitor Path Clearance and Provide Feedback.** During the adaptation stage, the robot continues to monitor whether the path to the end goal is clear (e.g., door is open wide enough, elevator door is open). If not, the robot prompts the user to continue the action if applicable ("*keep opening the door*"). Otherwise, the robot prompts the user for confirmation of proceeding ("*elevator door is open, press forward button to continue*").

## 5 User Study

We conducted a user study with the following objectives.

- **O1**: Compare the system's performance with established mobility aids (white cane and guide dog) in typical environmental interactions.
- **O2**: Examine how the dual-mode system design with adaptation improves users' experience in typical environmental interaction tasks.
- **O3**: Understand to what extent the system supports BVI users in completing navigation tasks safely and successfully, including both wayfinding and environmental interactions.

### 5.1 Participants

We recruited nine participants (P1-P9, 6 male, 3 female, **??**), aged between 50 and 75 years ($M = 62.6$), including eight white cane users and one guide dog user. These participants did not overlap with the formative interview participants. Six participants reported blindness acquired later in life, while three were blind from birth.

Data from the eight white cane users were used for the main analyses (Sec. 6.1–6.3). The guide dog user (P3) was analyzed separately in Sec. 6.4. Most participants traveled independently at least once per week. Each participant was compensated USD 38.

### 5.2 Experiment Design

We conducted the experiment in two parts.

**Part 1 (Task-level study).** Participants completed four environmental interaction tasks (*Elevator*, *Pull-Door*, *Push-Door*, *Chair*). For **O1**, *Full System* was compared with participants' habitual mobility aids (*White Cane* or *Guide Dog*), with an initial directional cue (e.g., "the elevator is at your 3 o'clock") provided to indicate the approximate target location. For **O2**, we compared *Full System* with a non-adaptive baseline representative of common assistive navigation systems [13, 35, 39]. The baseline stopped near the target and required users to complete the interaction independently, without adapting to user movement. The stopping position was selected to avoid nearby obstacles, not interfere with user actions, and maintain a fixed distance from the target for consistency. The baseline also omitted side-positioning announcements and user confirmation before switching back to lead mode.

We measured system performance with three metrics:

- **Efficiency:** Completion time (in seconds) coded from video recordings. For *Full System* and *Non-Adaptive*, voice instruction duration was excluded, as prompts serve as auxiliary cues and their length can be system-adjusted. For each task, we further divided the process into three sub-stages: (i) *Approach*—from the starting point until the robot stops in front of the target object; (ii) *Locate*—from the stop point until the user locates the operational element; and (iii) *Interact*—from the onset of interaction until the user and robot pass through and resume navigation.
- **Independence:** The number of experimenter interventions (triggered if the user was stuck for more than 5s).
- **Smoothness:** The number of unintended collisions or contacts between the user's body, cane, or robot and environmental elements (e.g., walls, doors, chairs).

After completing each task with all three solutions, participants rated their preferences across solutions (**Task Survey**) and the usefulness of *Full System*'s interactive features (**Feature Survey**) on a 5-point Likert scale, and provided open-ended feedback.

**Part 2 (Route-level study).** To address **O3**, participants completed a 5–10 minute navigation route from a bus stop to a laboratory office, including elevator use and spring-loaded doors. We mapped the route in advance with FAST-LIO [40].

We measured the same three system performance metrics (*efficiency, independence, smoothness*), with efficiency coded as overall completion time (without sub-stages). Participants also completed the NASA Task Load Index (TLX) [14] on a 7-point Likert scale and responded to open-ended questions about their overall experience.

### 5.3 Procedure
After informed consent, participants completed a brief practice session in a separate environment to familiarize themselves with the robot's leash cues, voice instructions, and compliant behaviors.

In **Part 1**, participants completed all four tasks using three navigation aids (*Full System*, *Non-Adaptive*, and *White Cane/Guide Dog*), with two trials per aid. Aid order was counterbalanced.

In **Part 2**, participants walked a predefined route using a *White Cane/Guide Dog* (with experimenter-provided instructions) and then retraced it with *Full System*. They subsequently completed the TLX and provided open-ended feedback.

The entire study lasted approximately 3.5 hours, and the tasks were video recorded for analysis.

## 6 Results & Discussion
In this section, we present results addressing our three objectives. For **O1**, we compare the system performance and user evaluation of *Full System* against the white cane (Sec. 6.1). For **O2**, we examine the effect of the dual-mode system by comparing *Full System* with *Non-Adaptive* (Sec. 6.2). Finally, for **O3**, we compare system performance along a complete route between *Full System* and the *White Cane/Guide Dog* (Sec. 6.3). In addition, Sec. 6.4 presents a comparison with an animal guide dog. All statistical analyses used one-way ANOVAs with Tukey-adjusted post-hoc comparisons where applicable.

### 6.1 Comparison to Traditional White Cane
We compared the performance and user experience **(N=8)** between *Full System* and the *White Cane* in four task scenarios in terms of efficiency, independence, smoothness, and user feedback.

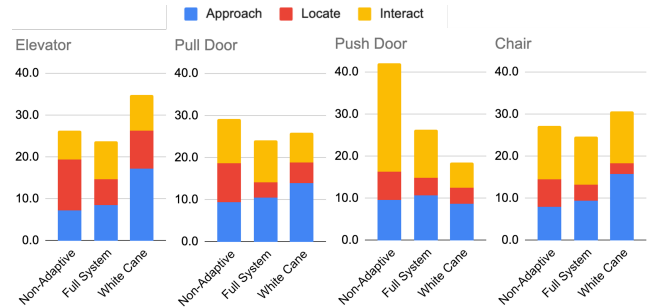

Figure 4: Time spent on each stage of the four tasks.

**Efficiency.** As shown in Table 1 and Figure 4, *Full System* completed three of the four tasks faster than the *White Cane*. The timing of users' arrival at the interaction targets suggests that *Full System*'s primary advantage over the *White Cane* lies in its precise navigation and localization capabilities, as reflected in the *Approach* and *Locate* stages. In the *Push-Door* task, *Full System* was slower, likely because the task required minimal navigation (following a hallway and pushing open the door). Statistical analyses showed a significant difference in overall task completion time for the elevator scenario (mean difference = 11.00 s, $p = .005$); differences in the other scenarios were not significant (see Appendix). This pattern likely reflects the difficulty of the elevator task: the elevator was in an unpredictable location within an open space, the call button could be on either side, and it was small and hard to locate without the robot's precise guidance.

**Independence.** As shown in Table 1, although interventions were generally rare, *Full System* required fewer than the *White Cane* in the *Elevator*, *Pull-Door*, and *Chair* tasks, aligning with the efficiency results.

**Smoothness.** Across all tasks, *Full System* had fewer collision cases compared to the *White Cane* (Table 1). The improvement was most pronounced in the *Elevator* and *Chair* tasks, where the environments were more cluttered.

**Task survey.** Users rated their overall experience and ease of use during different task steps on a five-point Likert scale (1 = strongly disagree, 5 = strongly agree; Appendix Figs. **??**–**??**). Compared with the *White Cane*, *Full System* reliably improved overall subjective experience across navigation scenarios (general experience: elevators, $F = 6.40$, $p = .006$, Tukey $p = .005$; doors, $F = 4.73$, $p = .026$, Tukey $p = .025$), with the largest benefits in target localization (elevators, $F = 21.32$, $p < .001$; elevator call buttons, $F = 9.35$, $p = .001$; chairs, $F = 15.38$, $p < .001$; all Tukey $p \leq .002$). No other measures showed reliable differences (Tukey $p > .05$).

**Interview results.** Participants emphasized that unstructured or open environments were particularly challenging with a white cane and valued the robot's ability to guide them precisely to interaction points. For example, in the chair task, scattered objects increased the cognitive load of identifying the correct target —"*I have to feel around to find the chair amid various objects. The robot simply brings me there*" (P6). In open spaces, users described greater spatial uncertainty, which the robot mitigated through direct guidance —"*If it is a big room I could be anywhere. The robot guide dog just brings me to the target, and I can trust it*" (P3). In contrast, tasks in more structured environments, such as approaching a door from a corridor, were perceived as easier with a cane —"*I just tap the wall and based on the sound I can locate the door fairly easily*" (P1).

Table 1: Average task performance across conditions (N=8, mean ± standard deviation; best values are in **bold**).

| Metric | Elevator | | | Pull-Door | | | Push-Door | | | Chair | | |
|---|---|---|---|---|---|---|---|---|---|---|---|---|
| | NoAdapt | Full | W-Cane | NoAdapt | Full | W-Cane | NoAdapt | Full | W-Cane | NoAdapt | Full | W-Cane |
| Time (s) | 26.2 ± 5.0 | **22.6 ± 6.0** | 33.6 ± 13.0 | 28.3 ± 9.6 | **23.7 ± 4.8** | 26.0 ± 10.0 | 42.0 ± 14.8 | 25.4 ± 5.3 | **18.5 ± 5.3** | 27.2 ± 8.1 | **24.6 ± 4.5** | 30.7 ± 9.6 |
| Interventions | 0.1 ± 0.3 | **0.0 ± 0.0** | 0.4 ± 0.6 | **0.0 ± 0.0** | **0.0 ± 0.0** | 0.3 ± 0.6 | 0.1 ± 0.3 | 0.1 ± 0.3 | **0.0 ± 0.0** | 0.2 ± 0.4 | **0.0 ± 0.0** | 0.2 ± 0.5 |
| Collisions (Hits) | **0.0 ± 0.0** | **0.0 ± 0.0** | 5.9 ± 7.2 | 0.1 ± 0.3 | 0.1 ± 0.4 | 1.4 ± 1.7 | 0.1 ± 0.3 | 0.1 ± 0.3 | 0.6 ± 0.8 | 0.2 ± 0.6 | **0.0 ± 0.0** | 2.0 ± 2.7 |

**Summary.** Taken together, the results show that the advantages of *Full System* over *White Cane* stem from accurate localization and pointing, thereby validating **D1**.

## 6.2 Comparison to Non-Adaptive System

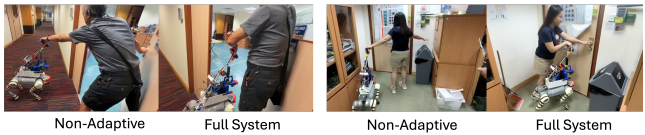

Figure 5: Comparison of the *Non-Adaptive* and *Full System*. Left: During *Interact* Stage, the *Full System* follows the user closely while avoiding obstacles. Right: During *Locate*, the *Full System* positions the user on the correct side and points accurately to the target.

We compared the performance and user experience **(N=8)** between *Full System* and *Non-Adaptive* in four task scenarios in terms of efficiency, independence, smoothness, and user feedback.

**Efficiency.** As shown in Figure 4, *Full System* completed all tasks faster than *Non-Adaptive*, with a significant advantage in the *Push-Door* task (mean difference = 16.6 s, $p < .001$, see Appendix for full results in all scenarios). The overall advantage was driven primarily by the *Locate* stage: *Full System* stopped close to the interaction target, oriented its head directly toward it, and positioned the user on the correct side, which made the target easier to reach (e.g., Figure 5 right). Although this fine-tuned positioning slightly increased the *Approach* stage, it consistently shortened the *Locate* stage across all tasks, yielding a net benefit. Tukey-adjusted post-hoc comparisons confirmed that the *Locate* stage was significantly shorter for *Full System* than for *Non-Adaptive* in every task ($p < .02$).

The overall improvement arises from the dual-mode design. Because the system cannot know in advance whether a door requires pulling or pushing, *Full System* adjusts its position dynamically as the interaction unfolds. This allows the robot to stop close to the target while avoiding obstacles (Figure 5). In contrast, *Non-Adaptive*—lacking adaptation—must halt at a safe distance to avoid blocking either case, often forcing users to overextend their reach between the leash and the handle. As a result, *Full System* provided more efficient and natural access to interaction targets, satisfying both accurate localization (**D1**) and unobtrusive positioning (**D2**).

**Task survey.** Participants consistently rated *Full System* higher than *Non-Adaptive*. On a 5-point Likert scale, mean differences were 1.00 for the door (Appendix Fig. 7), 0.88 for the elevator (Appendix Fig. 6), and 0.17 for the chair (Appendix Fig. 8). Differences were largest in the *Locate* stage (door 1.43, elevator 1.25, chair 0.50), highlighting *Full System*'s advantage for precisely guiding the user towards interaction targets. Statistical analyses showed that *Full System* provided additional benefits over *Non-Adaptive* mainly for interaction-critical steps: users found it easier to locate elevator call buttons ($F = 9.35$, $p = .001$; Tukey $p = .026$) and navigate push doors (opening: $F = 15.23$, $p < .001$; Tukey $p < .001$; passing through: $F = 8.55$, $p = .002$; Tukey $p = .008$). Other measures did not show significant differences.

**Feature survey.** Participants rated the usefulness of various features of *Full System* on a 5-point scale (1 = not useful at all, 5 = very useful; Appendix Fig. ??, ??, ??). Features included announcing which side to stand on ($M = 4.51$), announcing the precise location of the interaction target ($M = 4.71$), following the user's movement during operation ($M = 4.59$), and asking the user to confirm before moving ($M = 4.56$). On average, users gave high ratings (>4) to most features, confirming design requirements **D2** and **D3**.

**Interview results.** Participants emphasized that *Full System* substantially reduced their "searching time," which enhanced their sense of competence. For example, in the *Push-Door* task, several users described difficulty with *Non-Adaptive* due to its distant stopping position —"*I need to let go of the robot handle to push the door open, but it may be hard to find the handle*" (P5).

Interviews highlighted two key interactive features. First, the side announcement (Sec. 4.3.2) directed users which side to approach, making reaching motions more intuitive; without it, participants reported more hesitation —"*I need to think about which side to approach from, and I'm a bit concerned there could be obstacles on the side*" (P6). Second, the user confirmation (Sec. 4.3.3) ensured that users regained balance before the robot resumed guiding; without it, some participants found the automatic continuation of *Non-Adaptive* unsafe —"*It just goes without asking for my confirmation, and I almost lost balance when pulling the door*" (P1).

Notably, some participants preferred the simpler behavior of Non-Adaptive in more straightforward tasks (e.g., Chair, Pull-Door), finding it less mentally demanding: —"*With the earlier system [Full System], I feel more confident with finding buttons, but for chairs with a large back, the simpler one [Non-Adaptive] is already fine*" (P3). As participants gained experience with the system, some acted before the voice instructions finished, reflecting individual differences in preferred instruction detail and suggesting the need for adaptive, customizable communication in future designs.

**Summary.** In summary, the comparison between *Full System* and *Non-Adaptive* indicates that benefits arise from the *adaptation module*, supporting **D1** and **D2**.

## 6.3 Complete Route Performance

As shown in Table 2, *Full System* achieved shorter overall completion time than the *White Cane* ($M = 307.6$, $SD = 29.5$ vs. $M = 399.8$, $SD = 103.8$; $t(5) = -1.997$, $p = .102$), and significantly lower NASA–TLX workload ($M = 26.1$, $SD = 17.3$ vs. $M = 62.5$, $SD = 20.4$; $t(7) = -2.98$, $p = .020$). The route included straight and curved segments, heavy pedestrian traffic, as well as elevators, doors, and chairs, and much of the efficiency gain came from **unstructured** areas (e.g., a crowded canteen path), where cane users often required assistance —"*I would definitely need help if it's my first time traveling here*" (P7).

Table 2: Objective performance metrics and workload (NASA-TLX) in route-level experiment (N=8, mean ± standard deviation). Best values for each metric are in bold text.

| Metric | *Full System* | *White Cane* |
| --- | --- | --- |
| Time (s) | **307.57 ± 29.50** | 399.83 ± 103.83 |
| Intervention | **0.29 ± 0.49** | 2.00 ± 2.37 |
| Collision | **0.00 ± 0.00** | 0.67 ± 0.82 |
| NASA-TLX | **26.07 ± 17.31** | 62.54 ± 20.43 |

Shorter completion time and lower workload in *Full System* aligns with observations that cane users exhibited more deviations, interventions, and obstacle contacts. Overall, *Full System* required only **0.29** experimenter interventions on average. In one case (**P6**), a door was not detected under low evening light; the robot safely reverted to normal guidance and awaited help, which the user viewed positively. The results indicate *Full System*'s robustness in complex conditions with inconsistent lighting.

Participants also highlighted the natural transition between wayfinding and environmental interaction, and the smooth switching between floors. Overall, *Full System* was perceived as a practical and supportive aid that reduces physical and cognitive effort while maintaining reliable route-level performance and improving independence —"*I can definitely see this system being very useful in places like shopping malls, hospitals, and cafeterias. I rarely go to these places alone, but with this robot, it will be much easier*" (P4).

## 6.4 Comparison To Traditional Guide Dog

We conducted a comparative study (**N=1**) with an animal guide dog following the same protocol as the white cane experiment. Although only one guide dog user (P3) participated due to limited local adoption, the results reveal several interesting contrasts.

In Part 1, *Full System* unexpectedly outperformed the guide dog in efficiency, particularly in the elevator task (Guide Dog: 56 s vs. *Full System*: 24 s). This difference was largely due to the guide dog's lack of semantic understanding of the environment and its tendency to become distracted (e.g., wandering toward nearby pedestrians), resulting in significant detours. The user also had to repeat commands "*Find the elevator*" multiple times to keep the dog focused. Similarly, when locating a door, the guide dog occasionally missed or overshot the target, delaying the progress (Guide Dog: 37 s vs. *Full System*: 28 s). However, once the user found the interaction target, the guide dog coordinates smoothly with the user, likely due to its agility, soft body, and long-established rapport with the user.

At the route level (Part 2), the guide dog completed the journey faster than *Full System*, but required two user interventions due to distraction. This advantage is likely attributable to the route's largely straightforward path-following segments, where the guide dog's speed and agility exceed those of the robot.

Overall, although the guide dog is skilled in obstacle avoidance and coordination with its user, it is less efficient and reliable than *Full System* in unstructured or unfamiliar environments, particularly for precise wayfinding and locating interaction targets.

## 6.5 Overall Discussion

Our study demonstrated the effectiveness of our system across diverse environmental interaction tasks and full multi-level navigation routes, showing clear benefits over traditional aids. The advantages were most evident in tasks requiring precise localization of interaction targets, where the system drastically reduced the time needed to locate and operate objects. Improvements were observed not only in objective metrics (e.g., efficiency, independence, and smoothness), but also in user-reported usability, reflecting reduced stress and workload in navigating complex environments.

The dual-mode human-robot collaboration approach, which enables accurate pointing, effective communication, and dynamic adaptation to the user's movement, was a key factor in these improvements. By maintaining close proximity without obstructing the user and supporting task-aware guidance, *Full System* outperformed the non-adaptive baseline in terms of efficiency, smoothness, and user experience. Interestingly, some participants preferred the reduced movements of the non-adaptive system in scenarios where precise positioning was less critical (e.g., locating a chair back), indicating the potential for customizable guidance levels that adapt to task complexity and individual user preference.

Several limitations point to future work. Our study involved a limited number of participants and short-term evaluations; longitudinal deployments are needed to examine novelty effects, trust, and long-term adoption. In Part 2 study, route order was not counterbalanced due to logistical constraints. Although routes were reversed rather than repeated, full counterbalancing would be preferable in future studies. Future research should also investigate deployment in broader environments such as cafeterias with additional interactions (e.g., making payments by tapping).

Overall, our study establishes dual-mode guiding system as a realistic and highly valuable mobility aid for BVI users, extending existing solutions to support complex environmental interactions. Importantly, our approach highlights the value of involving end users early in the design process, ensuring the system meets real-world needs and supports meaningful interactions. More broadly, these results contribute to research on human–robot interaction by illustrating how adaptive role switching between robot initiative and user control can improve system performance, usability, and task success in assistive settings, opening new possibilities for collaborative guidance systems.

## Acknowledgments

This research is supported by the Ministry of Education, Singapore under an AcRF Tier 1 grant (No. 251RES2406). We thank Yuxuan Zhang and Peng Sun for their help with the user studies.